%% file: main.tex
\documentclass{sigchi2}

\usepackage{todonotes}

\usepackage{ifthen}

\makeatletter
\renewcommand\@cite[2]{%
Ref.~#1\ifthenelse{\boolean{@tempswa}}
{, \nolinebreak[3] #2}{}
}
\renewcommand\@biblabel[1]{[1]}
\makeatother

\usepackage{balance}       
\usepackage{graphics}      
\usepackage[T1]{fontenc}   
\usepackage{txfonts}
\usepackage{mathptmx}
\usepackage[pdflang={en-US},pdftex]{hyperref}
\usepackage{color}
\usepackage{booktabs}
\usepackage{textcomp}
\usepackage{xspace}
\usepackage{multirow}
\usepackage{microtype}        
\usepackage{ccicons}          

\usepackage{todonotes}

\usepackage{xcolor}

 \makeatletter
 \DeclareRobustCommand\onedot{\futurelet\@let@token\@onedot}
 \def\@onedot{\ifx\@let@token.\else.\null\fi\xspace}

 \def\etal{{et al}\onedot}
 \makeatother

\def\plaintitle{Sign Language Recognition, Generation, and Translation: \\An Interdisciplinary Perspective}

\def\emptyauthor{}
\def\plainkeywords{sign language; recognition, translation, and generation; ASL}

\makeatletter
\def\url@leostyle{%
  \@ifundefined{selectfont}{
    \def\UrlFont{\sf}
  }{
    \def\UrlFont{\small\bf\ttfamily}
  }}
\makeatother
\def\pprw{8.5in}
\def\pprh{11in}

\definecolor{linkColor}{RGB}{6,125,233}
\hypersetup{%
  pdftitle={\plaintitle},
  pdfauthor={\emptyauthor},
  pdfkeywords={\plainkeywords},
  pdfdisplaydoctitle=true, 
  bookmarksnumbered,
  pdfstartview={FitH},
  colorlinks,
  citecolor=black,
  filecolor=black,
  linkcolor=black,
  urlcolor=linkColor,
  breaklinks=true,
  hypertexnames=false
}

\newcommand{\slp}{sign language processing}
\newcommand{\Slp}{Sign language processing}

\usepackage[acronym,xindy,nonumberlist]{glossaries}
\input{acronyms}
\makeglossaries

\begin{document}
\pagenumbering{arabic}
\title{\plaintitle}

\numberofauthors{12}
\author{Danielle Bragg$^1$ \hspace{5mm} Oscar Koller$^2$ \hspace{5mm} Mary Bellard$^2$ \hspace{5mm} Larwan Berke$^3$ \\
Patrick Boudreault$^4$ \hspace{5mm} Annelies Braffort$^5$ \hspace{5mm} Naomi Caselli$^6$ \hspace{5mm} Matt Huenerfauth$^3$ \\
Hernisa Kacorri$^7$ \hspace{5mm} Tessa Verhoef$^8$ \hspace{5mm} Christian Vogler$^4$ \hspace{5mm} Meredith Ringel Morris$^1$ \\ \\
\normalfont{\large $^1$Microsoft Research - Cambridge, MA USA \& Redmond, WA USA \email{\{danielle.bragg,merrie\}@microsoft.com}} \\
\normalfont{\large $^2$Microsoft - Munich, Germany \& Redmond, WA USA \email{\{oscar.koller,marybel\}@microsoft.com}} \\
\normalfont{\large $^3$Rochester Institute of Technology - Rochester, NY USA \email{\{larwan.berke,matt.huenerfauth\}@rit.edu}} \\
\normalfont{\large $^4$Gallaudet University - Washington, DC USA \email{\{patrick.boudreault,christian.vogler\}@gallaudet.edu}} \\
\normalfont{\large $^5$LIMSI, CNRS, Paris-Saclay University - Orsay, France \email{annelies.braffort@limsi.fr}} \\
\normalfont{\large $^6$Boston University - Boston, MA USA \email{nkc@bu.edu}} \\
\normalfont{\large $^7$University of Maryland - College Park, MD USA \email{hernisa@umd.edu}} \\
\normalfont{\large $^8$Leiden University - Leiden, Netherlands \email{t.verhoef@liacs.leidenuniv.nl}} \\
}

\maketitle
\pagestyle{empty}

\input{abstract}






\input{introduction}
\input{rel_work}
\input{method}
\input{state_of_the_art}
\input{challenges}
\input{call_to_action}
\input{contributions}
\input{conclusion}

\section{Acknowledgments}

The authors would like to thank all workshop participants. We also thank Bill Thies for helpful discussions and prudent advice. This material is based on work supported by Microsoft, NIH R21-DC016104, and NSF awards 1462280, 1746056, 1749376, 1763569, 1822747, 1625793, and 1749384.

%
%
%
%
%
\balance{}

\bibliographystyle{SIGCHI-Reference-Format2}
\bibliography{bragg,koller,uiux}
\printglossary[type=\acronymtype]
\printglossary[type=main]
\end{document}

%% file: acronyms.tex
\newglossaryentry{aslg}{
    name={ASL},
    description={American Sign Language, the primary language of the Deaf community in the United States (and several other parts of the world)}}
\newacronym{asl}{ASL}{American Sign Language\glsadd{aslg}}

\newglossaryentry{cnng}{%
   name={CNN},%
   description={A Convolutional Neural Network is a specific type of neural network that makes use of efficient parameter sharing and has revolutionized computer vision}}
\newacronym{cnn}{CNN}{convolutional neural network\glsadd{cnng}}

\newglossaryentry{werg}{%
  name={Word Error Rate},%
  description={A standard evaluation criterion in speech recognition and sign language recognition, and is based on the edit distance between reference and hypothesis}} 

\newglossaryentry{gloss}{%
  name={gloss},%
  description={Transliteration of one language (e.g., \gls{aslg}) with words of another language (e.g., English). This retains the grammatical structure of the target language (e.g., \gls{aslg})}}

\newglossaryentry{hamnosys}{%
  name={HamNoSys},%
  description={The Hamburg Notation System~\cite{prillwitz_hamnosys_1989} is a phonetic transcription scheme that allows  explicit annotation of individual subcomponents of sign language such as handshapes, movements, etc}}

\newglossaryentry{epenthesis}{%
  name={epenthesis},%
  description={Insertion of \gls{phonological} features into or between signs, often in the form of movement for sign languages}}

\newglossaryentry{co-articulation}{%
  name={co-articulation},%
  description={The influence of surrounding signs/context on the production of the current sign, for example the ending of one sign affecting the beginning of the next}}

\newglossaryentry{vocabulary}{%
  plural={vocabularies},
  name={vocabulary},%
  description={The set of unique signs (or words) that occur in a language or dataset}
}

\newglossaryentry{signer-independent}{
  name={signer-independent},%
  description={Characterizes a realistic sign language recognition experimental setup where the signers that are tested on are explicitly excluded from the training set}}

\newglossaryentry{real-life}{
  name={real-life},%
  description={Describes sign language that has not been scripted or tailored to suit machine learning algorithms, but is rather produced naturally}}

\newglossaryentry{continuous}{
  name={continuous},%
  description={Refers to long phrases or full sentences as opposed to single, isolated signs (or words)}}

\newglossaryentry{depiction}{
  name={depiction},%
  description={Depiction refers to the ability to represent semantic information visually.  Enacting an event or using  \glspl{classifier} are examples of depiction. Depiction is structured and highly flexible and many aspects of depiction have no spoken language corollary}}

\newglossaryentry{uiuxg}{
  name={User Interface/User Experience},
  description={UI refers to user interface, while UX refers to user experience. Technology designers (and \gls{hci} researchers) are often trained to focus on these crucial aspects of technology}
}

\newglossaryentry{phonological}{
  name={phonological},
  description={Having to do with phonology, the smallest units of a language, and how they are put together. For spoken languages the components are sounds, and for sign languages they are handshapes, locations, and movements etc}
}

\newglossaryentry{lexical}{
  name={lexical},
  description={Having to do with lexicon, or the vocabulary}
}

\newglossaryentry{classifier}{
  name={classifier},
  description={A handshape used to represent a particular class of objects, actions, or ideas}
}

\newglossaryentry{fingerspelling}{
  name={fingerspelling},
  description={The use of a series of manual letters (handshapes representing letters), to spell out a written/spoken word in a sign language}
}

\newglossaryentry{audism}{
  name={audism},
  description={Oppression or discrimination of the Deaf minority by the hearing majority. This marginalization has a long history, and may be overt or covert}
}

\newglossaryentry{annotation}{
  name={annotation},
  description={Descriptions of sign language content, typically in a writing system, note-taking system, or other discrete system that a computer can read}
}

\newglossaryentry{deafculture}{
  name={Deaf culture},
  description={The culture of Deaf communities, of which sign languages are a central component. Capitalized ``Deaf'' refers to these cultures, while lowercase ``deaf'' refers to audiological status}
}

\newacronym{aam}{AAM}{active appearance model}
\newacronym{ai}{AI}{Artificial Intelligence}
\newacronym{am}{AM}{acoustic model}
\newacronym{ann}{ANN}{artificial neural network}
\newacronym{ap}{AP}{average precision}
\newacronym{ar}{AR}{action recognition}

\newacronym{asllrp}{ASLLRP}{American Sign Language Linguistic Research Project}
\newacronym{aslr}{ASLR}{automatic sign language recognition}
\newacronym{asr}{ASR}{automatic speech recognition}
\newacronym{au}{AU}{action unit}
\newacronym{auc}{AUC}{area under curve}
\newacronym{auslan}{Auslan}{Australian sign language}
\newacronym{avsr}{AVSR}{audio-visual speech recognition}
\newacronym{bcc}{BCC}{binary connected components analysis}
\newacronym{bfgs}{BFGS}{Broyden-Fletcher-Goldfarb-Shanno}
\newacronym{bhmm}{BHMM}{Bernoulli \protect\gls{hmm}}
\newacronym{bic}{BIC}{Bayesian information criterion}
\newacronym{blstm}{BLSTM}{bi-directional long short term memory}
\newacronym{bovw}{BOVW}{bag of visual words}
\newacronym{bow}{BOW}{bag of words}
\newacronym{bsl}{BSL}{British sign language}
\newacronym{cart}{CART}{classification and regression tree}
\newacronym{cbir}{CBIR}{content-based image retrieval}
\newacronym{cc}{CC}{classifier combination}
\newacronym{cer}{CER}{character error rate}
\newacronym{cmllr}{CMLLR}{constrained maximum likelihood linear regression}
\newacronym{cslr}{CSLR}{continuous sign language recognition}
\newacronym{ctc}{CTC}{connectionist temporal classification}
\newacronym{cv}{CV}{computer vision}
\newacronym{damp}{DAMP}{DARPA Arabic machine-print}
\newacronym{dbn}{DBN}{deep belief network}
\newacronym{dc}{DC}{direct current}
\newacronym{dcs}{DCS}{disparity cost slices}
\newacronym{dct}{DCT}{discrete cosine transform}
\newacronym{dgs}{DGS}{German sign language}
\newacronym{dhh}{DHH}{Deaf and hard of hearing}
\newacronym{dnn}{DNN}{deep neural network}
\newacronym{dog}{DoG}{difference of Gaussians}
\newacronym{dp}{DP}{dynamic programming}
\newacronym{dpf}{GPF}{dynamic partial function}
\newacronym{dpt}{DPT}{dynamic programming tracking}
\newacronym{dsgs}{DSGS}{Swiss German sign language}
\newacronym{dtw}{DTW}{dynamic time warping}
\newacronym{ebw}{EBW}{extended Baum Welch}
\newacronym{eer}{EER}{equal error rate}
\newacronym{emd}{EMD}{earth movers distance}
\newacronym{em}{EM}{expectation maximization}
\newacronym{er}{ER}{error rate}
\newacronym{facs}{FACS}{Facial Action Coding System}
\newacronym{fmllr}{fMLLR}{feature space \protect\glstext{mllr}}
\newacronym{fps}{fps}{frames per second}
\newacronym{fra}{FRA}{frame recognition accuracy}
\newacronym{fsa}{FSA}{finite state automaton}
\newacronym{fsw}{FSW}{Formal and Regular SignWriting}
\newacronym{gdl}{GDL}{glyph dependent length}
\newacronym{ghmm}{GHMM}{Gaussian \protect\gls{hmm}}
\newacronym{gmmhmm}{GMM-HMM}{Gaussian mixture model \protect\gls{hmm}}
\newacronym{gis}{GIS}{generalised iterative scaling}
\newacronym{gmd}{GMD}{Gaussian mixture densities}
\newacronym{gmm}{GMM}{Gaussian mixture model}
\newacronym{gnu}{GNU}{GNU's not Unix}
\newacronym{gpl}{GPL}{General Public License}
\newacronym{gtf}{GTF}{global texture features}
\newacronym{hci}{HCI}{Human-Computer Interaction}
\newacronym{hdm}{HDM}{histogram distortion model}
\newacronym{hltpr}{HLTPR}{Human Language Technology and Pattern Recognition}
\newacronym{hmm}{HMM}{hidden Markov model}
\newacronym{hog}{HoG}{histogram of oriented gradients}
\newacronym{hp}{HP}{hand position}
\newacronym{hsv}{HSV}{hue-saturation-value}
\newacronym{ht}{HT}{hand trajectory}
\newacronym{htk}{HTK}{Hidden Markov Model Toolkit}
\newacronym{hv}{HV}{hand velocity}
\newacronym{hwr}{HWR}{handwriting recognition}
\newacronym{iapr}{IAPR}{International Association for Pattern  Recognition}
\newacronym{icdar}{ICDAR}{International Conference on Document Analysis and Recognition}
\newacronym{icfhr}{ICFHR}{International Conference on Frontiers in Handwriting Recognition}
\newacronym{icr}{ICR}{intelligent character recognition}
\newacronym{idf}{IDF}{inverse document frequency}
\newacronym{idm}{IDM}{image distortion model}
\newacronym{ifh}{IFH}{invariant feature histogram}
\newacronym{ilsvrc}{ILSVRC}{ImageNet Large-Scale Visual Recognition Challenge}
\newacronym{imerr}{IER}{image error rate}
\newacronym{ip}{IP}{interest point}
\newacronym{irma}{IRMA}{Image Retrieval in Medical Applications}
\newacronym{is}{IS}{International Sign}
\newacronym{isl}{ISL}{Irish Sign Language}
\newacronym{iwr}{IWR}{intelligent word recognition}
\newacronym{iwslt}{IWSLT}{international workshop on spoken language  translation}
\newacronym{jsd}{JSD}{Jensen-Shannon divergence}
\newacronym{klt}{KLT}{Kanade-Lukas-Tomasi}
\newacronym{knn}{KNN}{$k$ nearest neighbor}
\newacronym{l1o}{L1O}{leaving one out}
\newacronym{lbfgs}{LBFGS}{limited memory \protect\gls{bfgs}}
\newacronym{lbg}{LBG}{Linde-Buzo-Gray}
\newacronym{lbp}{LBP}{local binary pattern}
\newacronym{lbw}{LBW}{Lancaster-Oslo-Bergen, Brown and Wellington}
\newacronym{lda}{LDA}{linear discriminant analysis}
\newacronym{ldc}{LDC}{Linguistic Data Consortium}
\newacronym{lf}{LF}{local feature}
\newacronym{lis}{LIS}{Italian sign language}
\newacronym{lm}{LM}{language model}
\newacronym{lob}{LOB}{Lancaster-Oslo-Bergen}
\newacronym{lsf}{LSF}{Langue des Signes Fran{\c}aise}
\newacronym{lstm}{LSTM}{long short term memory}
\newacronym{lvcsr}{LVCSR}{large vocabulary continuous speech recognition}
\newacronym{map}{MAP}{mean average  precision}
\newacronym{mce}{MCE}{minimum classification error}
\newacronym{mds}{MDS}{multi-dimensional scaling}
\newacronym{me}{ME}{maximum entropy}
\newacronym{mfdi}{MFDI}{mean face difference image}
\newacronym{mil}{MIL}{multiple instance learning}
\newacronym{mir}{MIR}{Mallinkrodt Institute of Radiology}
\newacronym{mit}{MIT}{Massachusetts Institute of Technology}
\newacronym{mle}{MLE}{model length estimation}
\newacronym{mllr}{MLLR}{maximum likelihood linear regression}
\newacronym{ml}{ML}{maximum likelihood}
\newacronym{mlpghmm}{MLP-GHMM}{\protect\gls{mlp}-\protect\gls{ghmm}}
\newacronym{mlp}{MLP}{multi-layer perceptron}
\newacronym{mmac}{MMAC}{Multi-Modal Arabic Corpus}
\newacronym{mmiconf}{MMI-conf}{confidence-based \protect\gls{mmi}}
\newacronym{mmi}{MMI}{maximum mutual information}
\newacronym{mmmiconf}{M-MMI-conf}{confidence-based \protect\gls{mmmi}}
\newacronym{mmmi}{M-MMI}{margin-based \protect\gls{mmi}}
\newacronym{mmpeconf}{M-MPE-conf}{confidence-based \protect\gls{mmpe}}
\newacronym{mmpe}{M-MPE}{margin-based \protect\gls{mpe}}
\newacronym{mnist}{MNIST}{modified \protect\gls{nist}}
\newacronym{mpeconf}{MPE-conf}{confidence-based \protect\gls{mpe}}
\newacronym{mpeg}{MPEG}{Moving Picture Experts Group}
\newacronym{mpe}{MPE}{minimum phone error}
\newacronym{mser}{MSER}{maximally stable extremal regions}
\newacronym{msrc}{MSRC}{Microsoft Research Cambridge}
\newacronym{mt}{MT}{machine translation}
\newacronym{mwe}{MWE}{minimum word error}
\newacronym{ngt}{NGT}{Dutch sign language}
\newacronym{nist}{NIST}{National Institute for Standards and Technology}
\newacronym{nlp}{NLP}{Natural Language Processing}
\newacronym{nn}{NN}{nearest neighbour}
\newacronym{numa}{NUMA}{non-uniform memory architecture}
\newacronym{oao}{OAO}{one against one}
\newacronym{oatr}{OATR}{one against the rest}
\newacronym{ocr}{OCR}{optical character recognition}
\newacronym{odel}{DEL}{object deletion}
\newacronym{oerr}{OER}{object error rate}
\newacronym{ohsu}{OHSU}{Oregon Health and Science University}
\newacronym{oins}{INS}{object insertion}
\newacronym{oov}{OOV}{out of vocabulary}
\newacronym{osub}{SUB}{object substitution}
\newacronym{p2dhmm}{P2DHMM}{pseudo two-dimensional \protect\gls{hmm}}
\newacronym{pascal}{PASCAL}{Pattern Analysis, Statistical Modelling and Computational Learning}
\newacronym{patdb}{PATDB}{Printed Arabic Text Database}
\newacronym{paw}{PAW}{piece of Arabic word}
\newacronym{pca}{PCA}{principal components analysis}
\newacronym{pc}{PC}{personal computer}
\newacronym{pdf}{PDF}{probablity density function}
\newacronym{pdm}{PDM}{point density model}
\newacronym{peir}{PEIR}{Pathology Education Instructional Resources}
\newacronym{poicaam}{POICAAM}{project-out inverse-compositional \protect\gls{aam}}
\newacronym{pp}{PP}{perplexity}
\newacronym{qbve}{QBVE}{query by visual example}
\newacronym{qs}{QS}{quotient of sums}
\newacronym{rampn}{RAMP-N}{\protect\Gls{rwth} Arabic Machine-Print Newspaper}
\newacronym{rasr}{RASR}{RWTH Aachen University Speech Recognizer}
\newacronym{rast}{RAST}{recognition by adaptive subdivision of the transformation space}
\newacronym{rbf}{RBF}{radial basis function}
\newacronym{relu}{ReLU}{rectified linear unit}
\newacronym{rgb}{RGB}{red green blue}
\newacronym{rimes}{RIMES}{Reconnaissance et Indexiation de donn\'ees Manuscrites et de fac simil\'ES}
\newacronym{rnn}{RNN}{recurrent neural network}
\newacronym{roc}{ROC}{receiver operating characteristic}
\newacronym{roi}{ROI}{region-of-interest}
\newacronym{rprop}{RProp}{resilient backpropagation}
\newacronym{rs}{RS}{relevance score}
\newacronym{rsv}{RSV}{retrieval status value}
\newacronym{rwth-asr}{RWTH ASR}{RWTH Aachen University Speech Recognition}
\newacronym{rwth-ocr}{RWTH OCR}{RWTH Aachen University Optical Character Recognition}
\newacronym{rwth}{RWTH}{RWTH Aachen University}
\newacronym{sat}{SAT}{speaker adaptive training}
\newacronym{see}{SEE}{signed exact English}
\newacronym{sgd}{SGD}{stochastic gradient descent}
\newacronym{sift}{SIFT}{scale invariant feature transformation}
\newacronym{signspeak}{SignSpeak}{SignSpeak}
\newacronym{slt}{SLT}{sign language translation}
\newacronym{slr}{SLR}{sign language recognition}
\newacronym{sl}{SL}{sign language}
\newacronym{smt}{SMT}{statistical machine translation}
\newacronym{ssd}{SSD}{sum of squared distances}
\newacronym{ssl}{SSL}{Swedish sign language}
\newacronym{surf}{SURF}{speeded-up robust features}
\newacronym{svd}{SVD}{singular value decomposition}
\newacronym{svm}{SVM}{support vector machine}
\newacronym{sv}{SV}{support vector}
\newacronym{swv}{SWV}{supervised writing variants}
\newacronym{tcstar}{TC-STAR}{Technology and Corpora for Speech to  Speech Translation}
\newacronym{tc}{TC}{technical committee}
\newacronym{tdps}{TDPs}{time distortion penalties}
\newacronym{tdp}{TDP}{time distortion penalty}
\newacronym{ter}{TER}{tracking error rate}
\newacronym{tet}{TET}{PDFlib Text Extraction Toolkit}
\newacronym{tfidf}{TF/IDF}{text frequency/inverse document frequency}
\newacronym{tf}{TF}{text frequency}
\newacronym{trap}{TRAP}{TempoRAl Pattern}
\newacronym{trec}{TReC}{Text Retrieval Conference}
\newacronym{ttr}{TTR}{types/token ratio}
\newacronym{uiux}{UI/UX}{\gls{uiuxg}}
\newacronym{uibk}{UIBK}{University of Innsbruck}
\newacronym{upv}{UPV}{Universidad Polit{\'e}cnica de Valencia}
\newacronym{url}{URL}{unique resource locator}
\newacronym{usps}{USPS}{US Postal Service}
\newacronym{uw}{UW}{University of Washington}
\newacronym{vjd}{VJD}{Viola \& Jones Detector}
\newacronym{vjt}{VJT}{Viola \& Jones Tracker}
\newacronym{vj}{VJ}{Viola \& Jones}
\newacronym{vm}{VM}{visual model}
\newacronym{voc}{VOC}{visual object classes challenge}
\newacronym{vsa}{VSA}{visual speaker adaptation}
\newacronym{vts}{VTS}{virtual training sample}
\newacronym{wat}{WAT}{writer adaptive training}
\newacronym{wer}{WER}{Word Error Rate\glsadd{werg}}
\newacronym{wta}{WTA}{winner-takes-all}
\newacronym{zow}{ZOW}{zero-order warping}


%% file: abstract.tex
\begin{abstract}

Developing successful sign language recognition, generation, and translation systems requires expertise in a wide range of fields, including computer vision, computer graphics, natural language processing, human-computer interaction, linguistics, and Deaf culture. 
Despite the need for deep interdisciplinary knowledge, existing research occurs in separate disciplinary silos, and tackles separate portions of the sign language processing pipeline. 
This leads to three key questions: 1) What does an interdisciplinary view of the current landscape reveal? 2) What are the biggest challenges facing the field? and 3) What are the calls to action for people working in the field?
To help answer these questions, we brought together a diverse group of experts for a two-day workshop. 
This paper presents the results of that interdisciplinary workshop, providing key background that is often overlooked by computer scientists, a review of the state-of-the-art, a set of pressing challenges, and a call to action for the research community.

\end{abstract}

%% file: introduction.tex
\section{Introduction}

Sign language recognition, generation, and translation is a research area with high potential impact. (For brevity, we refer to these three related topics as ``\slp'' throughout this paper.) 
According to the World Federation of the Deaf, there are over 300 sign languages used around the world, and 70 million deaf people using them \cite{WFD}. 
Sign languages, like all languages, are naturally evolved, highly structured systems governed by a set of linguistic rules.  They are distinct from spoken languages -- i.e., \gls{asl} is not a manual form of English -- and do not have standard written forms. 
However, the vast majority of communications technologies are designed to support spoken or written language (which excludes sign languages), and most hearing people do not know a sign language. As a result, many communication barriers exist for deaf sign language users.

\Slp{} would help break down these barriers for sign language users. 
These technologies would make voice-activated services newly accessible to deaf sign language users -- for example, enabling the use of personal assistants (e.g., Siri and Alexa) by training them to respond to people signing. 
They would also enable the use of text-based systems -- for example by translating signed content into written queries for a search engine, or automatically replacing displayed text with sign language videos. 
Other possibilities include automatic transcription of signed content, which would enable indexing and search of sign language videos, real-time interpreting when human interpreters are not available, and many educational tools and applications.



Current research in \slp{} occurs in disciplinary silos, and as a result does not address the problem comprehensively. 
For example, there are many computer science publications presenting algorithms for recognizing (and less frequently translating) signed content. 
The teams creating these algorithms often lack Deaf members with lived experience of the problems the technology could or should solve, and lack knowledge of the linguistic complexities of the language for which their algorithms must account. 
The algorithms are also often trained on datasets that do not reflect real-world use cases. 
As a result, such single-disciplinary approaches to \slp{} have limited real-world value \cite{erard2017sign}.

To overcome these problems, we argue for an interdisciplinary approach to \slp. 
Deaf studies must be included in order to understand the community that the technology is built to serve. 
Linguistics is essential for identifying the structures of sign languages that algorithms must handle. 
\Gls{nlp} and \gls{mt} provide powerful methods for modeling, analyzing, and translating. Computer vision is required for detecting signed content, and computer graphics are required for generating signed content. 
Finally, \gls{hci} and design are essential for creating end-to-end systems that meet the community's needs and integrate into people's lives.

This work addresses the following questions:

\begin{description}
    \item[Q1:] What is the current state of \slp, from an interdisciplinary perspective? 
    \item[Q2:] What are the biggest challenges facing the field, from an interdisciplinary perspective? 
    \item[Q3:] What calls to action are there for the field, that resonate across disciplines? 
\end{description}

To address these questions, we conducted an interdisciplinary workshop with 39 participants. The workshop brought together academics from diverse backgrounds to synthesize the state-of-the-art in disparate domains, discuss the biggest challenges facing \slp{} efforts, and formulate a call-to-action for the research community. This paper synthesizes the workshop findings, providing a comprehensive interdisciplinary foundation for future research in \slp. 
The audience for this paper includes both newcomers to \slp{} and experts on a portion of the technology seeking to expand their perspective. 

The main contributions of this work are:

\begin{itemize}
    \item orientation and insights for researchers in any domain, in particular those entering the field
    \item highlighting of needs and opportunities for interdisciplinary collaboration
    \item prioritization of important problems in the field for researchers to tackle next
\end{itemize}

%% file: rel_work.tex
\section{Background and Related Work}

Building successful sign language processing systems requires an understanding of \gls{deafculture} in order to create systems that align with user needs and desires, and of sign languages to build systems that account for their complex linguistic aspects. Here, we summarize this background, and we also discuss existing reviews of sign language processing, which do not take a comprehensive view of the problem.

\subsection{Deaf Culture}

Sign language users make up cultural minorities, united by common languages and life experience. 
Many people view deafness not as a disability, but as a cultural identity \cite{glickman1993deaf} with many advantages \cite{bauman2014deaf}. When capitalized, ``Deaf'' refers to this cultural identity, while lowercase ``deaf'' refers to audiological status. 
Like other cultures, \glspl{deafculture} are characterized by unique sets of norms for interacting and living. 
Sign languages are a central component of \glspl{deafculture}, their role in Deaf communities even characterized as sacred \cite{veditz}. 
Consequently, development of sign language processing systems is highly sensitive, and must do the language justice to gain adoption.  


Suppression of sign language communication has been a major form of oppression against the Deaf community. Such discrimination is an example of ``\gls{audism}'' \cite{bauman2004audism, eckert2013audism, humphries1975audism}. 
In 1880, an international congress of largely hearing educators of deaf students 
declared that spoken language should be used for educating deaf children, not sign language \cite{lane2017chronology}. Subsequently, oralism was widely enforced, resulting in training students to lip-read and speak, with varying success. 
Since then, Deaf communities have fought to use sign languages in schools, work, and public life (e.g., \cite{geers2017early}).  
Linguistic work has helped gain respect for sign languages, by establishing them as natural languages \cite{stokoe_dictionary_1965}. 
Legislation has also helped establish legal support for sign language education and use (e.g.,  \cite{assembly2006convention}). 
This historical struggle can make development of sign language software particularly sensitive in the Deaf community. 


\subsection{Sign Language Linguistics}

Just like spoken languages, sign languages are composed of building blocks, or \gls{phonological} features, put together under certain rules. 
The seminal linguistic analysis of a sign language (\gls{asl}) revealed that each sign has three main \gls{phonological} features: handshape, location on the body, and movement  \cite{stokoe_sign_1960}. 
More recent analyses of sign languages offer more sophisticated and detailed \gls{phonological} analyses \cite{brentari1996trilled, van2002phonological, sandler_sign_2006,brentari_sign_2018}. 
While \gls{phonological} features are not always meaningful (e.g., the bent index finger in the sign APPLE does not mean anything on its own), they \textit{can} be \cite{brentari_handshape_2011}. 
For example, in some cases the movement of the sign has a grammatical function. In particular, the direction of movement in verbs can indicate the subject and object of the sentence.  

\Glspl{classifier} 
represent classes of nouns and verbs -- e.g., one handshape in \gls{asl} is used for vehicles, another for flat objects, and others for grabbing objects of particular shapes. 
The vehicle handshape could be combined with a swerving upward movement to mean a vehicle swerving uphill, or a jittery straight movement for driving over gravel. Replacing the handshape could indicate a person walking instead. These handshapes, movements, and locations are not reserved exclusively for \glspl{classifier}, and can appear in other signs. Recognition software must differentiate between such usages.

\Gls{fingerspelling}, where a spoken/written word is spelled out using handshapes representing letters, is prevalent in many sign languages. For example, \gls{fingerspelling} is often used for the names of people or organizations taken from spoken language. Its execution is subject to a high degree of coarticulation, where handshapes change depending on the neighboring letters \cite{keane_coarticulation_2012}. 
Recognition software must be able to identify when a handshape is used for \gls{fingerspelling} vs. other functionalities. 

Sign languages are not entirely expressed with the hands; movement of the eyebrows, mouth, head, shoulders, and eye gaze can all be critical \cite{wilbur2000phonological, brentari2018representing}. 
For example, in \gls{asl}, raised eyebrows indicate an open-ended question, and furrowed eyebrows indicate a yes/no question.  Signs can also be modified by adding mouth movements -- e.g., executing the sign CUP with different mouth positions can indicate cup size. Sign languages also make extensive use of \gls{depiction}: using the body to depict an action (e.g., showing how one would fillet a fish), dialogue, or psychological events \cite{dudis2004depiction}. Subtle shifts in body positioning and eye gaze can be used to indicate a referent. 
Sign language recognition software must accurately detect these non-manual components.

There is great diversity in sign language execution, based on ethnicity, geographic region, age, gender, education, language proficiency, hearing status, etc. 
As in spoken language, different social and geographic communities use different varieties of sign languages (e.g., Black ASL is a distinct dialect of \gls{asl} \cite{mccaskill2011hidden}). 
Unlike spoken languages, sign languages contain enormous variance in fluency. 
Most deaf children are born to hearing parents, who may not know sign language when the child is born. Consequently, most deaf sign language users learn the language in late childhood or adulthood, typically resulting in lower fluency \cite{mayberry2018rethinking}. 
Sign language processing software must accurately model and detect this variety, which increases the amount and variety of training data required.

It is difficult to estimate sign language \gls{vocabulary} size. Existing \gls{asl}-to-English dictionaries contain 5-10k signs (e.g., \cite{signingsavvy}). 
However, they are not representative of true size, as they lack \glspl{classifier}, \glspl{depiction}, and other ways signs are modulated to add adjectives, adverbs, and nuanced meaning.

\subsection{Reviews}

Existing reviews of \slp{} are largely technical and out-of-date, written before the advent of deep learning. 
Most reviews focus on a specific subarea, such as the software and hardware used to recognize signs \cite{al2012review, vijay2012recent, joudaki2014vision, cooper2011sign}. Few reviews discuss multiple subfields of sign language technologies (e.g., recognition, translation, \textit{and} generation). 
In this work, we provide a broader perspective that highlights common needs (e.g., datasets), and applications that blend multiple technologies. 
Unlike past reviews, we also articulate a call to action for the community, helping to prioritize problems facing the field.

Existing reviews also incorporate limited perspectives outside of computer science. 
In particular, few reviews incorporate the linguistic, social, and design perspectives needed to build sign language systems with real-world use. 
Some reviews consider a related discipline (e.g., linguistics in \cite{cooper2011sign}), but do not consider the full spectrum of disciplines involved. 
This work integrates diverse interdisciplinary perspectives throughout, 
providing deeper insight into how technologies align with human experiences, the challenges facing the field, and  opportunities for collaboration.


There is already interest among researchers in various fields in applying their methods to sign language applications. In particular, some technical reviews of gesture recognition touch on sign language recognition as an application domain (e.g., \cite{wu1999vision, murthy2009review, suarez2012hand, khan2012hand}). These reviews focus on algorithms for detecting fingers, hands, and human gestures. However, by framing sign language recognition as an application area, they risk misrepresenting sign language recognition as a gesture recognition problem, ignoring the complexity of sign languages as well as the broader social context within which such systems must function. 
In this work, we provide linguistic and cultural context in conjunction with algorithms, to establish a more accurate representation of the space. 

%% file: method.tex
\section{Method}


To help answer our guiding questions, we convened a two-day workshop with leading experts in \slp{} and related fields. 
Many of these participants continued on to synthesize the workshop findings in this paper. 

\subsection{Participants}

A total of 39 workshop attendees were recruited from universities and schools (18) and a tech company (21).  Academic participants were based in departments spanning computer science, linguistics, education, psychology, and Deaf studies. Within computer science, specialists were present from computer vision, speech recognition, machine translation, machine learning, signal processing, natural language processing, computer graphics,  human-computer interaction, and accessibility. 
Attendees from the tech company had roles in research, engineering, and program/product management. 
The number of participants present varied slightly over the two days. 

Participants were demographically diverse:
\begin{itemize}
\item Nationality: Attendees were an international group, currently based in the U.S., Europe, and Asia. 
\item Experience: Career level ranged from recent college graduates through senior professors and executives. 
\item Gender: 25 male, 14 female
\item Audiological status: 29 hearing (6 with Deaf immediate family members), 10 \gls{dhh}
\end{itemize}

\subsection{Procedure}

The workshop activities were structured to facilitate progress toward our three guiding questions (current landscape, biggest challenges, and calls to action). Day 1 provided the necessary domain-specific background knowledge, and Day 2 addressed our three  questions as an interdisciplinary group. Interpreters and captioners were available throughout. 

\textit{Day 1: Sharing interdisciplinary domain knowledge.}

\begin{itemize}
\item \textbf{Domain Lectures:} A series of 45-minute talks, covering relevant domains and given by domain experts: Deaf culture, sign language linguistics, natural language processing, computer vision, computer graphics, and dataset curation. 
\item \textbf{Panel:} A 45-minute panel on Deaf users' experiences with, needs for, and concerns about technology, with a Deaf moderator and four Deaf panelists. 
\end{itemize}

\textit{Day 2: Discussing problems and mapping the path forward.}

\begin{itemize}
    \item \textbf{Breakout Sessions:} Participants divided into smaller groups (8-9/group) to discuss specific topics, for 3.5 hours.
    
    The topic areas, outlined by the organizers and voted on by participants, were:
    \begin{enumerate}
        \item Sign Language Datasets
        \item Sign Language Recognition \& Computer Vision 
        \item Sign Language Modeling \& \gls{nlp}
        \item Sign Language Avatars \& Computer Graphics 
        \item Sign Language Technology \gls{uiux} Design
    \end{enumerate}
    
    Each group focused on the following questions: 
    \begin{enumerate}
        \item What is the state-of-the-art in this area? 
        \item What are the biggest current challenges in this area? 
        \item What are possible solutions to these challenges? 
        \item What is your vision of the future for this domain? 
        \item What is your call to action for the community?
    \end{enumerate}

    \item \textbf{Breakout Presentations:} Each breakout group reported back on their topic, through a slide presentation 
    mixed with discussion with the larger group (about 20 minutes each).
    
\end{itemize}

In the following sections, we summarize 
the content generated through this workshop exercise, organized by our three guiding questions. 

%% file: state_of_the_art.tex
\section{Q1: What is the current landscape?}

In this section, we synthesize each group's formulation of the current state-of-the-art. 
We note that some topics overlap. In particular, data is central to progress on all fronts, so we start with a summary of the landscape in sign language datasets.

\subsection{Datasets}

\begin{table*}[tbp]
    \centering 
    \setlength{\tabcolsep}{1.7pt}
    \begin{tabular}{|l|r|r|c|r|c|c|}
         \hline
         \multicolumn{1}{|c}{\multirow{2}{1cm}{\centering\textbf{Dataset}}} & \multicolumn{1}{|c}{\multirow{2}{2cm}{\centering\textbf{\Gls{vocabulary}}}} & \multicolumn{1}{|c}{\multirow{2}{2cm}{\centering\textbf{Signers}}} & \multicolumn{1}{|c|}{\multirow{2}{2cm}{\centering\textbf{\Gls{signer-independent}}}} & \multirow{2}{1cm}{\centering\textbf{Videos}} & \multicolumn{1}{|c|}{\multirow{2}{2cm}{\centering\textbf{\Gls{continuous}}}} & \multicolumn{1}{|c|}{\multirow{2}{1cm}{\centering\textbf{\Gls{real-life}}}}  \\ &&&&&&\\ \hline
         Purdue RVL-SLLL ASL \cite{kak_purdue_2002} & 104 & 14 & no & 2,576 & yes & no  \\
         RWTH Boston 104 \cite{zahedi_continuous_2006} & 104 & 3 & no & 201 & yes & no  \\
         Video-Based CSL \cite{huang_videobased_2018} & 178 & 50 & no & 25,000 & yes & no  \\
         Signum \cite{vonagris_video_2007} & 465 & (24 train, 1 test) - 25  & yes & 15,075 & yes & no  \\
         MS-ASL \cite{joze_msasl_2018} & 1,000 & (165 train, 37 dev, 20 test) - 222 & yes & 25,513 & no & yes  \\
         RWTH Phoenix \cite{forster_extensions_2014} & 1,081 & 9 & no & 6,841 & yes & yes  \\
         RWTH Phoenix SI5 \cite{koller2017:re-sign} & 1,081 & (8 train, 1 test) - 9 & yes & 4,667 & yes & yes  \\
         Devisign \cite{chai_devisign_2014} & 2,000 & 8 & no & 24,000 & no & no  \\
         \hline
    \end{tabular}
    \caption{Popular public corpora of sign language video. These datasets are commonly used for sign language recognition.}
    \vspace{-1em}
    \label{tab:datasets}
\end{table*}

Existing sign language datasets typically consist of videos of people signing. 
Video format can vary, and is often dependent on the recording device. For example, video cameras often produce MP4, OGG, or AVI format (among others). Motion-capture datasets have been curated, often by attaching sensors to a signer (e.g., \cite{lu2010collecting, heloir2005captured, berret2016collecting}). These datasets can be pulled from to generate signing avatars, and are often curated for this purpose. Depth cameras can also be used to capture 3D positioning. For example, the Kinect includes a depth sensor and has been used to capture sign language data (e.g., \cite{slrGroup, cooper_sign_2012, oszust2013polish}). Table~\ref{tab:datasets} summarizes public sign language video corpora commonly used for sign language recognition. (See \cite{konrad2012sign}'s survey for a more complete list of datasets, many of which are intended for linguistic research and education.)

The data collection method impacts both content and signer identity. 
For example, some corpora are formed of professional interpreters paid to interpret spoken content, such as news channels that provide interpreting \cite{forster_extensions_2014, koller2017:re-sign, camgoz_neural_2018}. 
Others are formed of expert signers paid to sign desired corpus content (e.g., \cite{kak_purdue_2002, zahedi_continuous_2006, vonagris_video_2007}). 
Yet other corpora consist sign language videos posted on sites such as YouTube (e.g., \cite{joze_msasl_2018}) -- these posters may be fluent signers, interpreters, or sign language students; such videos are typically of ``real-life'' signs (i.e., self-generated rather than prompted). 
The geographic region where the data is collected also dictates which sign language is captured. 
Many datasets have been curated by researchers and startups in the U.S., where \gls{asl} is the primary language of the Deaf community, and consequently contain \gls{asl}. Fewer datasets have been curated of other sign languages, though some exist (e.g., \cite{matthes2012dicta}). The \gls{vocabulary} size of these datasets varies from about 100--2,000 distinct signs (see Table~\ref{tab:datasets}).

\Glspl{annotation} may accompany video corpora. These \glspl{annotation} can demarcate components of signs (e.g., handshapes and movements), the identity and ordering of the signs, or a translation into another language like English. These \glspl{annotation} can take various formats, including linguistic notation systems (for sign components), English gloss (for sign identity and order), and English text (for translations). The \glspl{annotation} can be aligned at various levels of granularity. For example, the start and end of a handshape could be labeled, or the start and end of a full sentence. 
Generating \glspl{annotation} can be very time-intensive and expensive. \Gls{annotation} software has been developed to facilitate annotating videos, and is often used by linguists studying the language (e.g., ELAN \cite{elan} and Anvil \cite{anvil}). 
Because sign languages do not have a standard written form, large text corpora do not exist independent of videos. 

\subsection{Recognition \& Computer Vision}

Glove-based approaches to sign language recognition have been used to circumvent computer vision problems involved in recognizing signs from video. 
The first known work dates back to 1983, with a patent describing an electronic glove that recognized \gls{asl} fingerspellings based on a hardwired circuit ~\cite{grimes_digital_1983}. 
Since then, many other systems have been built for ``intrusive sign recognition,'' where signers are required to wear gloves, clothing, or other sensors to facilitate recognition (e.g., \cite{charayaphan_image_1992,fels_glovetalk_1993,liang_realtime_1998,oz_american_2011, cooper2011sign}).
%

Non-intrusive vision-based sign language recognition is the current dominant approach. Such systems minimize inconvenience to the signer (and, unlike gloves, have the potential to incorporate non-manual aspects of signing), but introduce complex computer vision problems. 
The first such work dates back to 1988, when Tamura~\etal \cite{tamura_recognition_1988} built a system to recognize 10 isolated signs of Japanese Sign Language using skin color thresholding. 
As in that seminal work, many other systems focus on identifying individual signs (e.g., \cite{grobel1997isolated, lang2012sign}).

Real-world translation typically requires continuous sign language recognition \cite{starner_realtime_1995, forster_modality_2013}, where a continuous stream of signing is deciphered. 
Continuous recognition is a significantly more challenging and realistic problem than recognizing individual signs, confounded by \gls{epenthesis} effects (insertion of extra features into signs), \gls{co-articulation} (the ending of one sign affecting the start of the next), and spontaneous sign production (which may include slang, non-uniform speed, etc.). 
%

To address the three-dimensionality of signs, some vision-based approaches use depth
cameras~\cite{uebersax_realtime_2011, zafrulla_american_2011}, multiple cameras~\cite{brashear_using_2003} or triangulation for 3D reconstruction~\cite{schmidt_using_2013,schmidt_enhancing_2013}. Some use colored gloves to ease hand and finger tracking~\cite{cooper_sign_2010}.
Recent advances in machine learning -- i.e., deep learning and \glspl{cnn} -- have improved  state-of-the-art computer vision approaches~\cite{koller_deep_2018}, though lack of sufficient training data currently limits the use of modern \gls{ai} techniques in this problem space.

Automatic recognition systems are transitioning from small, artificial \glspl{vocabulary} and tasks to larger real-world ones. Realistic scenarios are still very challenging for state-of-the-art algorithms. As such, 
recognition systems achieve only up to 42.8\% letter accuracy~\cite{shi_american_2018a} on a recently released  
real-life \gls{fingerspelling} dataset. 
A real-life continuous sign language video dataset has also been released~\cite{forster_extensions_2014}, and is used as a community benchmark. 
Given utterance- or sentence-level segmentation, recognition systems can reliably identify sign boundaries \cite{koller_weakly_2019}. 
For such challenging datasets (still only covering a \gls{vocabulary} of around 1,000 different signs), recognition algorithms can achieve a \gls{wer} of 22.9\%~\cite{cui_deep_2019} when trained and tested on the same signers, and a \gls{wer} of 39.6\% when trained and tested on different sets of signers. 

\subsection{Modeling \& Natural Language Processing}


Because sign languages are minority languages lacking data, the vast majority of work in \gls{mt} and \gls{nlp} focuses on spoken and written languages, not sign languages. 
While recognition handles the problem of identifying words or signs from complex signals (audio or video), \gls{mt} and \gls{nlp} typically address problems of processing language that has already been identified. 
These methods expect annotated data as input, which for spoken languages is commonly text (e.g., books, newspapers, or scraped text from the internet). 
Translation between spoken and signed languages (and vice versa) also typically requires intermediary representations of the languages that are computationally compatible. 

Various notation systems are used for computational modeling. 
Human-generated \glspl{annotation} are often in \gls{gloss}, a form of transliteration where written words in another language (e.g., English) are used to represents signs (e.g., \gls{asl}). 
Other writing systems have also been developed for people to use, including si5s \cite{clark_how_2012} and SignWriting \cite{sutton_sign_2000}. 
Separate notation systems have been developed for computers to represent sign languages during modeling and computation;  
\gls{hamnosys} \cite{hanke2004hamnosys} is one of the most popular, designed to capture detailed human movements and body positioning for computer modeling.  To facilitate structured storage, XML-based markup languages have also been developed, e.g. Signing Gesture Markup Language (SiGML) \cite{elliott2000development}, which is compatible with \gls{hamnosys}.

Sign language translation systems can either use predefined intermediary representations of the languages involved, or learn their own representations (which may not be human-understandable). 
Methods that use predefined representations are highly compatible with grammatical translation rules 
(e.g., \cite{davydov2017information, zhao2000machine, elliott2000development, veale1998challenges}). 
Methods that do not use such representations typically use some form of deep learning or neural networks, which learn model features (i.e. internal representations) that suit the problem. These methods have been used for recognition combined with translation, processing complete written sentences in parallel with signed sentences \cite{koller_deep_2018, koller16:hybridsign, camgoz_neural_2018}. Such techniques are often used in computer vision systems, and overlap with works presented in the previous section. 


\subsection{Avatars \& Computer Graphics}

Sign language avatars (computer animations of humans) can provide content in sign language, making information accessible to \gls{dhh} individuals who prefer sign language or have lower literacy in written language~\cite{huenerfauth2009sign}. Because sign languages are not typically written, these videos can be preferable to text. Videos of human signers or artist-produced animations provide similar viewing experiences, but avatars are more appropriate when automatic generation is desirable (e.g., for a website with unstable content). 
Current pipelines typically generate avatars based on a symbolic representation of the signed content prepared by a human author (e.g. \cite{al2018modeling, adamo2015aslpro, braffort2016kazoo, vcom3d, bangham2000virtual}). 
When the avatar is generated as part of a translation system (e.g., \cite{karpouzis2007educational, elliott2008linguistic}), an initial translation step converts spoken/written language into a symbolic representation of the sign language (as described in the previous section).  
Whether human-authored or automatically translated, a symbolic plan is needed for the sign-language message. While multiple representations have been proposed (e.g. \cite{elliott2000development, braffort2016kazoo,adamo2015aslpro}), there is no universal standard.

Beginning with this symbolic plan, pipelines generating avatars typically involve a series of complex steps (e.g., as outlined in \cite{huenerfauth2009sign, gibet2011signcom}). 
Animations for individual signs are often pulled from lexicons of individual signs.  These motion-plans for the individual signs are produced in one of several ways: key-frame animations (e.g. \cite{huenerfauth2014release}), symbolic encoding of sub-sign elements (e.g. \cite{ebling2016}), or motion-capture recordings (e.g., \cite{segouat2009toward, gibet2011signcom}). 
Similarly, non-manual signals 
are pulled from complementary datasets (e.g., \cite{huenerfauth2006generating}) or synthesized from models (e.g., \cite{kacorri-huenerfauth-2016-continuous}). These elements are combined to create an initial motion script of the content. Next, various parameters (e.g., speed, timing) are set by a human, set by a rule-based approach (e.g. \cite{ebling2016}), or predicted via a trained machine-learning model (e.g., \cite{al2018modeling, lu2010collecting}). Finally, computer animation software renders the animation based on this detailed movement plan.

The state-of-the-art in avatar generation is not fully automated; all parts of current pipelines currently require human intervention to generate smooth, coherent signing avatars.  Prior research has measured the quality of avatar animations via perceptual and comprehension studies with \gls{dhh} participants \cite{huenerfauth2007evaluating}, including methodological research \cite{kacorri2017regression} and shared resources for conducting evaluation studies \cite{huenerfauth2014release}.

\subsection{\gls{uiux} Design}

The state-of-the-art of sign language output in user interfaces primarily centers around systems that use sign language video or animation content (e.g., computer-generated human avatars) to display information content.  
(Surveys of older work appear in \cite{huenerfauth2009sign} and \cite{al2018modeling}.) 
These projects include systems that provide sign language animation content on webpages to supplement text content for users.   There has also been some work on providing on-demand definitions of terminology in \gls{asl} (e.g., by linking to \gls{asl} dictionary resources \cite{hariharan2018evaluation}). 
As discussed in \cite{hariharan2018evaluation}, prior work has found that displaying static images of signs provides limited benefit, and generally users have preferred interfaces that combine both text and sign content.

Research on designing interactive systems with sign language recognition technologies has primarily investigated how to create useful applications despite the limited accuracy and coverage of current technology for this task.  
This has often included research on tools for students learning \gls{asl}, either young children (e.g., \cite{zafrulla2011copycat}) or older students who are provided with feedback as to whether their signing is accurate \cite{huenerfauth2017evaluation}.   
While there have been various short-lived projects and specific industry efforts to create tools that can recognize full phrases of \gls{asl} to provide communication assistance, few systems are robust enough for real-world deployment or use.  

%% file: challenges.tex
\section{Q2: What are the field's biggest challenges?}

In this section, we summarize the major challenges facing the field, identified by the interdisciplinary breakout groups. 

\subsection{Datasets}

Public sign language datasets have shortcomings that limit the power and generalizability of systems trained on them.

\begin{table*}[tbp]
    \centering
    \begin{tabular}{|l|l|l|}
         \hline
         & \multicolumn{1}{c|}{\textbf{Sign Language}} & \multicolumn{1}{c|}{\textbf{Speech}}   \\  \hline
         Modality & visual-gestural & aural-oral \\
         Articulators & manual, non-manual & vocal tract \\
         Seriality & low & high \\
         Simultaneity & high & low \\
         Iconicity & high & low \\
         \hline
         Task & recognition, generation, translation & recognition, generation, translation \\
         Typical articulated corpus size & <100,000 signs & 5 million words \\
         Typical annotated corpus size & <100,000 signs & 1 billion words \\
         Typical corpus vocabulary size & 1,500 signs & 300,000 words \\
         What is being modelled & 1,500 whole signs & 1,500 tri-phonemes \\
         Typical corpus number of speakers & 10 & 1,000 \\
         \hline
    \end{tabular}
    \caption{Comparison of sign language vs. speech datasets. Existing sign language corpora are orders of magnitude smaller than speech corpora. Because sign languages are not typically written, parallel written corpora do not exist for sign languages, as they do for spoken (and written) languages.}
    \vspace{-1em}
    \label{tab:sl_speech}
\end{table*}





\textit{Size:} Modern, data-driven machine learning techniques work best in data-rich scenarios. 
Success in speech recognition, which in many ways is analogous to sign recognition, has been made possible by training on corpora containing millions of words. 
In contrast, sign language corpora, which are needed to fuel the development of sign language recognition, are several orders of magnitude smaller, typically containing fewer than 100,000 articulated signs. (See Table \ref{tab:sl_speech} for a comparison between speech and sign language datasets.)

\textit{\Gls{continuous} Signing:} Many existing sign language datasets contain individual signs. Isolated sign training data may be important for certain scenarios (i.e., creating a sign language dictionary), but most real-world use cases of \slp{} involve natural conversational with complete sentences and longer utterances. 

\textit{Native Signers:} Many datasets allow novices (i.e., students) to contribute, or contain data scraped from online sources (e.g., YouTube \cite{joze_msasl_2018}) where signer provenance and skill is unknown. 
Professional interpreters, who are highly skilled but are often not native signers, are also used in many datasets (e.g., \cite{forster_rwthphoenixweather_2012}). The act of interpreting also changes the execution (e.g., by simplifying the style and vocabulary, or signing slower for understandability).  
Datasets of native signers are needed to build models that reflect this core user group.

\textit{Signer Variety:} The small size of current signing datasets and over-reliance on content from interpreters mean that current datasets typically lack signer variety. 
To accurately reflect the signing population and realistic recognition scenarios, datasets should include signers that vary by: gender, age, clothing, geography, culture, skin tone, body proportions, disability, fluency, background scenery, lighting conditions, camera quality, and camera angles. It is also crucial to have \gls{signer-independent} datasets, which allow people to assess generalizability by training and testing on different signers. 
Datasets must also be generated for different sign languages (i.e., in addition to \gls{asl}). 

\subsection{Recognition \& Computer Vision}

Despite the large improvements in recent years, there are still many important and unsolved recognition problems, which hinder real-world applicability. 

\textit{\Gls{depiction}:} \Gls{depiction} refers to visually representing or enacting content in sign languages (see Background \& Related Work), and poses unique challenges for recognition and translation. 
Understanding depiction requires exposure to \gls{deafculture} and linguistics, which the communities driving  progress in computer vision generally lack. 
Sign recognition algorithms are often based on speech recognition, which does not handle depictions (which are uncommon and unimportant in speech). As a result, current techniques cannot handle depictions. 
It is also difficult to create \gls{depiction} \glspl{annotation}. 
Countless depictions can express the same concept, and \gls{annotation} systems do not have a standard way to encode this richness. 


\textit{Annotations:} 
Producing sign language \glspl{annotation}, the machine-readable inputs needed for supervised training of \gls{ai} models, is time consuming and error prone. 
There is no standardized \gls{annotation} system or level of \gls{annotation} granularity. 
As a result, researchers are prevented from combining annotated datasets to increase power, and must handle low inter-annotator agreement. 
Annotators must also be trained extensively to reach sufficient proficiency in the desired \gls{annotation} system. Training is expensive, and constrains the set of people who can provide \glspl{annotation} beyond the already restricted set of fluent signers. 
The lack of a standard written form also prevents learning from naturally generated text -- e.g., \gls{nlp} methods that expect text input, using parallel text corpora to learn corresponding grammar and vocabulary, and more generally leveraging ubiquitous text resources. 

\textit{Generalization:} Generalization to unseen situations and individuals is a major difficulty of machine learning, and sign language recognition is no exception. 
Larger, more diverse datasets are essential for training generalizable models. We outlined key characteristics of such datasets in the prior section on dataset challenges.  
However, generating such datasets can be extremely time-consuming and expensive.


\subsection{Modeling \& Natural Language Processing}

The main challenge facing modeling and \gls{nlp} is the inability to apply powerful methods used for spoken/written languages due to language structure differences and lack of \glspl{annotation}. 

    \textit{Structural Complexity:} Many \gls{mt} and \gls{nlp} methods were developed for spoken/written languages. However, sign languages have a number of structural differences from these languages. These differences mean that straightforward application of \gls{mt} and \gls{nlp} methods will fail to capture some aspects of sign languages or simply not work. In particular, many methods assume that one word or concept is executed at a time. However, many sign languages are multi-channel, for instance conveying an object and its description simultaneously. Many methods also assume that context does not change the word being uttered; however, in sign languages, content can be spatially organized and interpretation directly dependent on that spatial context.
    
    \textit{Annotations:} Lack of reliable, large-scale \glspl{annotation} are a barrier to applying powerful MT and NLP methods to sign languages. 
    These methods typically take \glspl{annotation} as input, commonly text. 
    Because sign languages do not have a standard written form or a standard \gls{annotation} form, we do not have large-scale \glspl{annotation} to feed these methods. Lack of large-scale annotated data is similarly a problem for training recognition systems, as described in the previous section.

\subsection{Avatars \& Computer Graphics}

Avatar generation faces a number of technical challenges in creating avatars that are acceptable to Deaf users (i.e., pleasing to view, easy to understand, representative of the Deaf community, etc.). Some of these problems may be addressed by including Deaf people in the generation process \cite{kipp2011assessing}.

    \textit{Uncanny Valley:} Sign language avatars are subject to an uncanny valley \cite{mori2012uncanny}. 
    Avatars that are either very cartoonish or very human-like are fairly pleasing, but in-between can be disconcerting. 
    For example, in addition to providing semantically meaningful non-manual cues (e.g., raised eyebrows indicating a question), avatars must also have varied, natural facial expressions (i.e., not a robotic, stoic expression throughout). 
    It can be difficult to design avatars that fall outside of this valley.
    
    \textit{Realistic Transitions:} To illustrate why transitions between signs are difficult, consider a generation system that pulls from motion-capture libraries. The system can pull complete sign executions, but must then piece together these executions. One sign might end with the hands in one position, while the subsequent sign starts with the hands in another position, and the software must create a smooth, natural transition between. 
    
    \textit{Modeling Modulations:} In some sign languages, adjectives and adverbs are executed by modulating a noun or verb. For example, in \gls{asl}, PLANE RIDE is executed by moving a certain handshape through the air. BUMPY PLANE RIDE is identical, but with the movement made bumpy. Infinite such descriptors can be executed, and capturing them all in a motion-capture database is infeasible. Acceptable abstractions have not been standardized (e.g., in a writing system), so it is unclear how much real-life variation avatars must portray. 
    
    \textit{Finding Model Holes:} It is difficult to find holes in generation models, because the language space is large and rich, and the number of ways that signs can be combined in sequence grows exponentially. Testing all grammatical structures empirically 
    is not scalable. This ``unknown unknown'' problem is common to other machine learning areas (e.g., speech recognition \cite{hermansky2013multistream}). 
    
    \textit{Public Motion-Capture Datasets} Many motion-capture datasets used for avatar generation are owned by particular companies or research groups. Because they are not publicly available, research in this area can be impeded.


\subsection{\gls{uiux} Design}

Sign language \gls{uiux} design is currently confounded by technical limitations that require carefully scoped projects, many potential use cases requiring different solutions, and design choices that may have powerful social ramifications.

    \textit{Technical Limitations} A long-term goal in this space is full universal design of conversational agents. For example, if a system supports speech-based or text chat interaction, then it should also support input and output in sign language. However, given the current limitations of the component technologies, it may be useful for researchers to focus on more near-term research aims: for instance, if we have a sign language recognition system capable of recognizing some finite number of signs or phrases, then what types of applications can be supported within this limit (for different \gls{vocabulary} sizes)?  
    
    \textit{Varied Use Cases:} There are a huge number of use cases for \slp, requiring different interface designs. 
    For example, sign language recognition could be useful for placing a meal order in a drive-through restaurant, or for commanding a personal assistant. 
    Similarly, sign language generation may be used in various situations. For people who want to create websites that present signed content, avatars may be the most reasonable solution, as they allow for ease in editability, creation from anywhere, and scalability (cost).  However, people also want websites to be searchable and indexable, and videos and animations are difficult for current text-based search engines to index and search. 
    Combining text and video introduces layout problems, especially when text is automatically replaced with video. 
    These situations, and many others, have drastically different design criteria. 
    
    \textit{Language and Dialect Choice:} Many different sign languages exist, with many dialects for each. Choosing which one(s) a system will recognize or portray is a difficult problem with societal implications.  Minorities within Deaf communities may be further marginalized if their dialects are not represented. Similarly, failure to represent other types of diversity -- e.g., gender, race, education level, etc. -- could also be detrimental.
    

%% file: call_to_action.tex
\section{Q3: What are the calls to action?}

In this section, we outline an interdisciplinary call to action for the research community working on any piece of the end-to-end \slp{} pipeline. 
Once stated, these calls to action may seem intuitive, but have not previously been articulated, and have until now been largely disregarded. 

\subsection{Deaf Involvement}

In developing \slp, Deaf community involvement is essential at all levels, 
in order to design systems that actually match user needs, are usable, and to facilitate adoption of the technology. 
An all-hearing team lacks the lived experience of Deafness, and is removed from the use cases and contexts within which sign language software must function. 
Even hearing people with strong ties to the Deaf community are not in a position to speak for Deaf needs. Additionally, because of their perceived expertise in Deaf matters, they are especially susceptible to being involved in Deaf-hearing power imbalances. 
People who do not know a sign language also typically make incorrect assumptions about sign languages -- e.g., assuming that a particular gesture always translates to a particular spoken/written word. 
As a result, all-hearing teams are ill-equipped to design software that will be truly useful. 

It is also important to recognize individual and community freedoms in adopting technology. Pushing a technology can lead to community resentment, as in the case of cochlear implants for many members of sign language communities \cite{sparrow2005defending}. 
Disrespecting the Deaf community's ownership over sign languages also furthers a history of \gls{audism} and exclusion, which can result in the Deaf community rejecting the technology. 
For these reasons, a number of systems built by hearing teams to serve the Deaf community have failed or receive mixed reception (e.g., sign language gloves \cite{erard2017sign}). 

Deaf contributors are essential at \textit{every step} of research and development.
For example, involvement in the creation, evaluation, and ownership of sign language datasets is paramount to creating high-quality data that accurately represents the community, can address meaningful problems, and avoids cultural appropriation. 
Future datasets might take cultural competency into account by 1) being open-source and publicly available, 2) providing cultural context for challenges to ensure that computer vision experts competing on algorithmic performance understand the nature, complexity, and history of sign languages, and/or 3) providing more appropriate metrics developed by the Deaf community, beyond the current standard of \gls{wer}. 
Similarly, Deaf community involvement is fundamental to the creation of appropriate computational models, interface design, and overall systems. 

The importance of Deaf involvement is heightened by technology's impact on language. 
Written English is evolving right now with new spellings based on technological constraints like character limits on Twitter, and difficulty typing long phrases on phone keyboards. 
It is possible that signers would similarly adapt sign languages to better suit the constraints of computing technologies. For example, people might simplify vocabulary to aid recognition software, constrict range of motion to fit the technical limits of video communications \cite{keating2008}, or abstract away richness to support standardized writing or \gls{annotation}.

\fbox{\begin{minipage}{\columnwidth}
    \textbf{Call 1: Involve Deaf team members throughout.}\vspace{.5em}
    
    \textit{Deaf involvement and leadership are crucial for designing systems that are useful to users, respecting Deaf ownership of sign languages, and securing adoption.}
\end{minipage}}

\subsection{Application Domain}

There are many different application domains for \slp. Situations where an interpreter would be beneficial but is not available are one class of applications. This includes any point of sale, restaurant service, and daily spontaneous interactions (for instance with a landlord, colleagues, or strangers). 
Developing personal assistant technologies that can respond to sign language is another compelling application area. 
Each of these scenarios requires different solutions. Furthermore, these different use cases impose unique constraints on every part of the pipeline, including the content, format, and size of training data, the properties of algorithms, as well as the interface design. 
Successful systems require buy-in from the Deaf community, so ensuring that solutions handle application domains appropriately is essential.

Technical limitations impact which domains are appropriate to tackle in the near-term, and inform intermediary goals that which will ultimately inform end-to-end systems. 
Many of these intermediary goals are worth pursuing in and of themselves, and offer bootstrapping benefits toward longer-term goals. For example, a comprehensive, accurate sign language dictionary that lets users look up individual signs would be an important resource for sign language users and learners alike, and would also inform model design for continuous sign language recognition. 
In addition, support for everyday use of sign language writing would make text-based resources accessible to sign language users in their language of choice, and would also organically generate an annotated corpus of sign language that could be used to learn language structure. 

\fbox{\begin{minipage}{\columnwidth}
    \textbf{Call 2: Focus on real-world applications.}\vspace{.5em}
    
    \textit{\Slp{} is appropriate for specific domains, and the technology has limitations. Datasets, algorithms, interfaces, and overall systems should be built to serve real-world use cases, and account for real-world constraints.}

\end{minipage}}

\subsection{Interface Design} 

The field currently lacks fundamental research on how users interact with sign language technology. A number of systems have been developed explicitly serving sign language users (e.g., messaging services \cite{fiveApp, ssms}, games \cite{lee2005gesture, brashear2006american}, educational tools \cite{reis2015asl, alshammari2018building}, webpages \cite{petrie2004augmenting}, dictionaries \cite{slinto, bragg2015user}, and writing support \cite{bianchini2012swift, bragg2018designing}). 
However, accompanying user studies typically focus on evaluating a single system, and do not outline principles of interaction that apply across systems. 
As a result, each team developing a new system must design their interface largely from scratch, uninformed by general design guidelines based on research.

Since many technologies required for end-to-end sign language translation are under development, it may be necessary for researchers to use Wizard-of-Oz style testing procedures (e.g., \cite{dahlback1993wizard}) to better understand how Deaf users would react to various types of user-interface designs. Recent work has used such approaches. For instance, researchers have used Wizard-of-Oz methodologies to study how Deaf users would like to issue commands to personal assistants \cite{gallaudet2019csun} or how Deaf users may benefit from a tool that enables \gls{asl} dictionary lookup on-demand when reading English text webpages \cite{hariharan2018evaluation}. 

Returning to the personal assistant application mentioned above, a Wizard-of-Oz methodology could be used to investigate interaction questions, such as how the user might ``wake up'' the system so it expects a command, and how the system might visually acknowledge a signed command (e.g., by presenting written-language text onscreen) and provide a response to the user (e.g., as written-language text or as sign-language animation).  Additionally, such simulations may also be used to determine how good these technologies must be before they are acceptable to users, i.e., what threshold of recognition accuracy is acceptable to users in specific use cases. Such work can set an agenda for researchers investigating the development of core sign-language recognition or synthesis technologies.

\fbox{\begin{minipage}{\columnwidth}
    \textbf{Call 3: Develop user-interface guidelines for sign language systems.}\vspace{.5em}
    
    \textit{Because \slp{} is still developing, we lack a systematic understanding of how people interact with it. Guidelines and error metrics for effective system design would support the creation of consistently effective interfaces.}
\end{minipage}}

\subsection{Datasets}

As highlighted throughout this work, few large-scale, publicly available sign language corpora exist. Moreover, the largest public datasets are orders of magnitude smaller than those of comparable fields like speech recognition. 
The lack of large-scale public datasets 
shifts the focus from algorithmic and system development to data curation. 
Establishing large, appropriate corpora would expedite technical innovation. 

In particular, the field would benefit from a larger body of research 
involving reproducible tasks. 
Publicly available data and competitive evaluations are needed to create
interest, direct research towards the challenges that matter (tackling depiction, generalizing to unseen signers, real-life data), and increase momentum. 
Furthermore, having open-source implementations of full pipelines would also foster faster adoption. 

There are four main approaches for collecting signing data, each of which has strengths and weaknesses. Developing multiple public data resources that span these four approaches may be necessary in order to balance these tradeoffs.
\begin{enumerate}
    \item Scraping video sites (e.g., YouTube) has many potential benefits: low cost, rapid collection of many videos, the naturalistic nature of the data, and potential diversity of participants. Its pitfalls include: privacy and consent of people in the videos, variability in signing quality, and lack of accompanying \glspl{annotation}. 
    \item Crowdsourcing data through existing platforms (e.g., Amazon Mechanical Turk) or customized sites (e.g., \cite{bragg2015user}) offers potential cost savings (particularly if participants contribute data for free), and the ability to reach diverse contributors (i.e., by removing geographic constraints). However, crowdsourcing is subject to quality control issues. In paid systems people may rush or ``cheat'' to earn more money, and in unpaid learning activities, well-intentioned learners may submit low-quality or incorrect data. 
    \item Bootstrapping, where products are released with limitations and gather data during use, is common to other \gls{ai} domains (e.g., voice recognition \cite{schalkwyk2010your}).
    This approach is cheap, collects highly naturalistic data, and may scale well. However, privacy and informed consent are potential pitfalls, and there is a cold-start problem -- can a useful application be created from current datasets to support this bootstrapping process, and can it achieve a critical mass of users?
    \item 
    In-lab collection allows for customized, high-end equipment such as high-resolution, high-frame-rate cameras, multiple cameras, depth-cameras, and motion-capture suits. 
    However, this type of controlled collection may result in less naturalistic content, higher costs that limit scalability, and lower participant diversity due to geographic constraints. 
    Models trained on such high-quality data also may not generalize to users with low-quality phone or laptop cameras. 
\end{enumerate}

Some metadata impacting data utility can only be gathered at the time of capture. 
In particular, demographics may be important for understanding biases and generalizability of systems trained on the data \cite{gebru2018datasheets}. 
Key demographics include signing fluency, language acquisition age, education (level, Deaf vs. mainstream), audiological status, socioeconomic status, gender, race/ethnicity, and geography. Such metadata can also benefit linguistics, Deaf studies, and other disciplines. 

Metadata regarding the data collection process itself (i.e., details enabling replication) are also vital to include so that others can add to the dataset. For example, if a dataset is gathered in the U.S., researchers in other countries could replicate the collection method to increase geographic diversity.

\fbox{\begin{minipage}{\columnwidth}
    \textbf{Call 4: Create larger, more representative, public video datasets.}\vspace{.5em}
    
    \textit{Large datasets with diverse signers are essential for training software to perform well for diverse users. Public availability is important for spurring developments, and for ensuring that the Deaf community has equal ownership.}
\end{minipage}}

\subsection{Annotations}


A standard \gls{annotation} system would expedite development of \slp. 
Datasets annotated with the standard system could easily be combined and shared. 
Software systems built to be compatible with that \gls{annotation} system would then have much more training data at their disposal. 
A standard system would also reduce \gls{annotation} cost and errors. 
As described earlier, the lack of standardization results in expensive training (and re-training) of annotators, and ambiguous, error-prone \glspl{annotation}. 

Designing the \gls{annotation} system to be appropriate for everyday reading and writing, or developing a separate standard writing system, would provide addition benefits. 
With such a system, email clients, text editors, and search engines would become newly usable in sign languages without translating into a spoken/written language. 
As they write, 
users would also produce a large annotated sign language corpus of naturally generated content, which could be used to better train models. 
However, establishing a standard writing system requires the Deaf community to reach consensus on how much of the live language may be abstracted away. Any writing system loses some of the live language (i.e., a transcript of a live speech in English loses pitch, speed, intonation, and emotional expression). Sign languages will be no different.

Computer-aided \gls{annotation} software has been proposed (e.g., \cite{crasborn_transcription_2015, dreuw2008towards, chetelat2008sign}), but could provide increased support due to recent advances in deep learning applied to sign language recognition. 
Current sign language modeling techniques could be used to aid the \gls{annotation} process in terms of both segmenting and transcribing the input video. 
Aided \gls{annotation} should leverage advances in modeling whole signs and also sign subunits~\cite{koller_deep_2016,koller_weakly_2019}. 
Annotation support tools could also alleviate problems with annotating depictions, as they could propose \glspl{annotation} conditioned on the translation and hence circumvent the problem of detailing the iconic nature of these concepts.

\fbox{\begin{minipage}{\columnwidth}
    \textbf{Call 5: Standardize the \gls{annotation} system and develop software for \gls{annotation} support.}\vspace{.5em}
    
    \textit{Annotations are essential to training recognition systems, providing inputs to \gls{nlp} and \gls{mt} software, and generating signing avatars. Standardization would support data sharing, expand software compatibility, and help control quality. Annotation support would help improve accuracy, reliability, and cost.}
\end{minipage}}

%% file: contributions.tex
\section{Contributions}

This paper provides an interdisciplinary perspective on the field of \slp. 
For computer scientists and technologists, it provides key background on Deaf culture and sign language linguistics that is often lacking, and contextualizes relevant subdomains they may be working within (i.e., \gls{hci}, computer vision, computer graphics, \gls{mt}, \gls{nlp}). 
For readers outside of computer science, it provides an overview of how \slp{} works, and helps to explain the challenges that current technologies face. 

In synthesizing the state-of-the-art from an interdisciplinary perspective, this paper provides orientation for researchers in any domain, in particular those entering the field. 
Unlike disciplinary reviews that focus on relevant work in a particular domain, we relate these domains to one another, and show how \slp{} is dependent on all of them. 

In summarizing the current challenges, this work highlights opportunities for interdisciplinary collaboration. Many of the problems facing the field cross disciplines.
In particular, questions of how to create datasets, algorithms, user interfaces, and a standard annotation system that meet technical requirements, reflect linguistics of the language, and are accepted by the Deaf community are large, open problems that will require strong, interdisciplinary teams. 

Finally, in articulating a call to action, this work helps researchers prioritize efforts to focus on the most pressing and important problems. 
Lack of data (in particular large, annotated, representative, public datasets) is arguably the biggest obstacle currently facing the field. 
This problem is confounded by the relatively small pool of potential contributors, recording requirements, and lack of standardized annotations. Because data collection is difficult and costly, companies and research groups are also incentivised to keep data proprietary. 
Without sufficient data, system performance will be limited and unlikely to meet the Deaf community's standards. 

Our workshop methodology used to provide this interdisciplinary perspective on \slp{} can be used as a model for other similarly siloed fields. 
While the general structure of the workshop is directly duplicable, some work would need to be done to tailor it to other fields (e.g., identifying relevant domains and domain experts).

%% file: conclusion.tex
\section{Conclusion}

In this paper, we provide an interdisciplinary overview of sign language recognition, generation, and translation. 
Past work on \slp{} has largely been conducted by experts in different domains separately, limiting real-world utility. 
In this work, we assess the field from an interdisciplinary perspective, tackling three questions: 1) What does an interdisciplinary view of the current landscape reveal? 2) What are the biggest challenges facing the field? and 3) What are the calls to action for people working in the field?

To address these questions, we ran an interdisciplinary workshop with 39 domain experts with diverse backgrounds. 
This paper presents the interdisciplinary workshop's findings, providing key background for computer scientists on Deaf culture and sign language linguistics that is often overlooked, a review of the state-of-the-art, a set of pressing challenges, and a call to action for the research community. 
In doing so, this paper serves to orient readers both within and outside of computer science to the field, highlights opportunities for interdisciplinary collaborations, and helps the research community prioritize which problems to tackle next (data, data, data!).